\documentclass{article}

     \usepackage[preprint,nonatbib]{neurips_2020}

\usepackage[utf8]{inputenc} 
\usepackage[T1]{fontenc}    
\usepackage{hyperref}       
\usepackage{url}            
\usepackage{booktabs}       
\usepackage{amsfonts}       
\usepackage{nicefrac}       
\usepackage{microtype}      

\usepackage{adjustbox}
\usepackage{microtype}
\usepackage{graphicx}
\usepackage{subfigure}
\usepackage{booktabs} 
\usepackage{amsmath,amssymb,amsfonts}
\usepackage{algorithmic}
\usepackage{graphicx}
\usepackage{textcomp}
\usepackage{xcolor}
\usepackage{booktabs}
\usepackage{multirow}
\usepackage{hyperref}
\usepackage{cleveref}
\usepackage{caption}
\usepackage{times}
\usepackage{fancyhdr}
\usepackage[ruled,vlined]{algorithm2e}
\include{pythonlisting}

\title{A Gradient-based Bilevel Optimization Approach for Tuning Hyperparameters in Machine Learning}

\author{%
  Ankur Sinha\\
  Production and Quantitative Methods\\
  Indian Institute of Management Ahmedabad\\
  Gujarat, India 380015 \\
  \texttt{asinha@iima.ac.in} \\
   \And
    Tanmay Khandait \\
    Production and Quantitative Methods\\
    Indian Institute of Management Ahmedabad\\
  Gujarat, India 380015 \\
  \texttt{tanmayk@iima.ac.in} \\
   \AND
    Raja Mohanty \\
    Production and Quantitative Methods\\
    Indian Institute of Management Ahmedabad\\
    Gujarat, India 380015 \\
  \texttt{rajam@iima.ac.in} \\
}

\def\reals{\mathbb{R}}

\def\comp{\raise 1pt \hbox{$\scriptstyle\circ$}}

\def\argmin{\mathop{\rm argmin}\limits}

\def\minimize{\mathop{\rm min}\limits}

\def\st{\mathop{\rm subject\ to}}

\def\upto{{\raise 1pt \hbox{$\scriptstyle \,\nearrow\,$}}}
\def\downto{{\raise 1pt \hbox{$\scriptstyle \,\searrow\,$}}}

\def\tos{\rightrightarrows}

\newcommand{\norm}[1]{\left\lVert#1\right\rVert}

\newtheorem{definition}{Definition}

\begin{document}

\maketitle

\begin{abstract}
  Hyperparameter tuning is an important and active area of research in machine learning, where the aim is to identify the optimal hyperparameters that provide the best performance on the validation set. Hyperparameter tuning is often achieved using naive techniques, such as random search and grid search. However, most of these methods seldom lead to an optimal set of hyperparameters and often get very expensive. The hyperparameter optimization problem is inherently a bilevel optimization task in nature, and there exist past studies that have attempted bilevel solution methodologies for solving this problem. In this paper, we propose a bilevel solution method for solving the hyperparameter optimization problem that does not suffer from the drawbacks of the earlier studies. The proposed method is general and can be easily applied to any class of machine learning algorithms. The idea is based on the approximation of the lower level optimal value function mapping, which is an important mapping in bilevel optimization and helps in reducing the bilevel problem to a single level constrained optimization task. The single-level constrained optimization problem is solved using the augmented Lagrangian method. We discuss the theory behind the proposed algorithm and perform extensive computational study on two datasets that confirm the efficiency of the proposed method. We perform a comparative study against grid search, random search and Bayesian optimization techniques that shows that the proposed algorithm is multiple times faster on problems with one or two hyperparameters. The computational gain is expected to be significantly higher as the number of hyperparameters increase. Corresponding to a given hyperparameter most of the techniques in the literature often assume a unique optimal parameter set that minimizes loss on the training set. Such an assumption is often violated by deep learning architectures and the proposed method does not require any such assumption.
\end{abstract}

\section{Introduction}\label{introduction}

In machine learning problems, finding the optimal hyperparameters has always been a cumbersome task. In essence, hyperparameters are those parameters that are external to the model and can't be learned using training data alone. The model specifications include network architecture (e.g., number of layers and nodes), optimization parameters (e.g., learning rate and momentum) and regularization (weight decay and dropout). The process of finding a set of optimal hyperparameters of a model, which would minimize the validation set loss is known as hyperparameter optimization. Finely tuned hyperparameters often lead to a high accuracy of a machine learning model, particularly when applied to data that the model might not have seen during the training phase.

The hyperparameter optimization problem is inherently a bilevel optimization task because of its hierarchical nature. The outer problem requires minimizing the validation set loss, with respect to some hyperparameters, and the inner problem requires minimizing the training set loss, with respect to the model parameters. Modeling the hyperparameter optimization problem as a bilevel optimization task is not new in the field of machine learning \cite{bennett08}. 
The area of bilevel optimization itself has two roots of origin; firstly, in the domain of game theory \cite{von1952theory}, where it is studied as a leader-follower problem or Stackelberg problem; and secondly, in the domain of mathematical programming \cite{bracken1973mathematical}, where these problems are studied as nested optimization problems or bilevel optimization problem. A number of review papers and books have been written in this area for which the readers may refer to \cite{sinha2017review,dempe2002foundations,bard2013practical}, among others. 



A significant body of literature exists on hyperparameter optimization; however, studies using a bilevel approach to solve this problem are relatively few. Naive techniques such as grid search and random search, are the most widely used methods for hyperparameter tuning. Here, the model is trained over a set of hyperparameters that are chosen, either randomly or on a grid, and the hyperparameter that minimizes the validation loss is accepted as the final choice \cite{bergstra2011algorithms}. \cite{bergstra2012random} showed that random search is more effective than grid search, and should be a preferred choice over grid search. An extension of random search, referred to as the Hyperband algorithm \cite{li2017hyperband}, uses a multi-armed bandit technique for hyperparameter configuration evaluations at each iteration and then allocating computational resources to promising configurations. Furthermore, Bayesian optimization, a model-based search method, has been the gold standard for hyperparameter optimization \cite{hutter2011sequential,bergstra2011algorithms,snoek2012practical,snoek2015scalable}. In Bayesian optimization, for every training run, one creates a probabilistic model for the objective function, which is called a surrogate function. Then by maximizing the acquisition function, which balances the trade-off between exploration and exploitation based on the posterior, an informed decision is made for the next hyperparameter value to train on. Commonly used methods for the probabilistic modelling of the objective function is Gaussian process \cite{snoek2012practical}, tree Parzen estimator \cite{bergstra2011algorithms} and sequential model-based optimization for algorithm configuration  \cite{hutter2011sequential}. For acquisition function, the most commonly used method is Expected Improvement Criterion \cite{bergstra2011algorithms}. 

Another branch of hyperparameter optimization which has gained a lot of traction in recent years is gradient based hyperparameter optimization, some aspects of our study would fall under this category, where the validation set loss with respect to hyperparameters is minimized using a gradient descent algorithm. The studies under this can be classified into two categories. The first category is an iterative approach where the best response function is approximated after some steps of gradient descent on the loss function \cite{maclaurin2015gradient,franceschi2018bilevel}, while another approach uses the implicit function theory to derive the hyper-gradients \cite{bengio2000gradient,pedregosa2016hyperparameter}. Recently, \cite{lorraine2018stochastic,mackay2019self} used the trust region method to locally approximate the best response function. Some of these methods involve approximating Hessians or its inverses, which is computationally very expensive. In our proposed approach, we formulate the hyperparameter optimization problem as a bilevel optimization task, and then approximate the lower level value function mapping, which is an important mapping that reduces the bilevel optimization problem into a single level constrained optimization problem. The single level constrained optimization problem is converted into an unconstrained problem using the augmented Lagrangian method, which we then solve using a gradient based method. One of the important features of our method is that it does not require computation or approximation of the Hessian or its inverse, which makes our algorithm applicable for hyperparameter tuning of deep learning architectures. Further, in deep learning architectures, it is common to have multiple (global) optimal solutions that minimize loss on the training set. Many of the methods fail to search on this set of multiple optimal solutions. The proposed method does not suffer from this drawback, which in our opinion, is a very important advantage over other methods in the literature.

The paper is organized as follows. To begin with, we provide a general formulation of the bilevel optimization problem in Section~\ref{bilevelhyper}, followed by the proposed approach in Section~\ref{propmethod}. 
Section~\ref{sec:experiments} provides the results from detailed experimentation on a regression and a classification problem, Section~\ref{sec:conclusions} provides the conclusions.

\begin{table}[]
\centering

\caption{Central Notation}
\label{tab:centralNotation}
\resizebox{\textwidth}{!}{%
\begin{tabular}{@{}lll@{}}
\toprule
\textbf{Category}                   & \textbf{Notation} & \multicolumn{1}{c}{\textbf{Description}} \\ \midrule
\multirow{4}{*}{Examples}           & \multirow{2}{*}{$S_{T}$}       &   \begin{tabular}[c]{@{}l@{}}$S_{T}= \{(x_i,y_i)\}_{i = 1}^{N^T}$; training set, where $x$ and $y$ are combination of \\input features and output classes  $N^T$ is the number of training examples\end{tabular}\\\cmidrule{2-3}
                                    &  \multirow{2}{*}{$S_{V}$}        & \begin{tabular}[c]{@{}l@{}} $S_{V}= \{(x_i,y_i)\}_{i = 1}^{N^V}$; validation set, where $x$ and $y$ are combination of \\input features and output classes  $N^V$ is the number of validation examples\end{tabular}\\ \cmidrule{1-3}
\multirow{2}{*}{Decision variables} & $\lambda \in \mathbf{R}^{n}$          & hyperparameter (upper level decision) \\\cmidrule{2-3}
                                    & $ w \in \mathbf{R}^{m}$               & model parameters (lower level decision)                      \\ \cmidrule{1-3}
Loss function                       & $l$                 &  average cross-entropy loss function                   \\ \cmidrule{1-3}
Regularization                      & $\Theta$                  &  L2 regularization                 \\\cmidrule{1-3}
\multirow{2}{*}{Objectives}         & $F$                 & $F(w) = l(w; S^V)$; validation set loss function (upper level objective)                  \\ \cmidrule{2-3}
                                    & $f$                 &$ f(\lambda, w) = l(w; S^{T}) + \Theta(w,\lambda)$; training set regularized loss function (lower level objective)                      \\\cmidrule{1-3}
Reaction set                     &  $\Psi: \mathbf{R}^{n} \rightarrow \mathbf{R}^{m}$                 &  $\Psi(\lambda)$ represents the optimal solution(s) of the lower level function for any upper level decision vector \\\cmidrule{1-3}
Optimal value function              &  $\varphi: \mathbf{R}^{n} \rightarrow \mathbf{R}$                 & represents the minimum lower level function value corresponding to an upper level decision vector                     \\ \bottomrule
\end{tabular}
}
\end{table}

\section{Bilevel Hyperparameter Optimization}\label{bilevelhyper}
As mentioned in Section~\ref{introduction}, in the context of hyperparameter optimization, the lower level problem minimizes the training set loss and the upper level problem minimizes the validation set loss. The loss on the training set is the function of model parameters and hyperparameters and the loss on validation set is the function of only model parameters. Note that the bilevel optimization structure imposes the necessary restriction that only the model parameters that minimize training set loss, are the feasible model parameters to be considered for the validation set loss evaluation. The notations used in our study are summarized in Table \ref{tab:centralNotation}.


\begin{definition}\label{def:bilevel1}
 Let $F: \mathbb{R}^{n} \rightarrow \mathbb{R}$ and $f: \mathbb{R}^{n} \times \mathbb{R}^{m} \rightarrow \mathbb{R}$ be the upper level (validation loss) and lower level (training loss) problems, respectively. The upper level objective function in this formulation depends on the lower level vector $w$, and the lower level objective function depends on the upper level vector $\lambda$ and lower level vector $w$. The lower level problem is, therefore, parameterized with respect to $\lambda$, and is optimized with respect to $w$.
\begin{align}\label{mod:originalBilevel}
\begin{split}
\minimize_{\lambda,w} \quad & F(w) \\
\st\quad  & \\
 & \hspace{-12mm} w \in \argmin_w \{f(\lambda,w)\}
 \end{split}
\end{align}
where, $w \in \mathbb{R}^{m}$ denotes the model parameters, and $\lambda \in \mathbb{R}^{n}$ denotes the hyperparameters.
\end{definition}

In case of multiple optimal solutions at the lower level, it is imperative to provide a rule for the treatment of solutions at the upper level. In our study, we will be following the optimistic formulation, wherein, in the presence of multiple lower level solutions, the best solution with respect to upper level objective gets chosen. This does not lead to any loss of generality for the hyperparameter optimization problem.

In our paper, we use L2 regularization (Tikhonov) term at the lower level, which has a parameter $\lambda$. In our experiments, we have also considered weighted L2 regularization, when $\lambda$ is a vector. The L2 regularization hyperparameter controls the penalty on the sum of squares of the model parameters (weights), thereby encouraging them to be smaller in magnitude. Therefore, the lower level objective is the following regularized training loss. 
\begin{align*}
    f(\lambda, w) = l(w; S^{T}) + \Theta(w,\lambda)
\end{align*}
where $S_{T}$ is the set of training examples, $l$ is the mean squared error or the average cross-entropy loss function, and $\Theta$ is the L2 regularization function defined as $\Theta = \lambda \norm{w}^2$. 
The upper level objective is the average validation loss, which is given as follows:
\begin{align*}
    F(w) = l(w; S^V)
\end{align*}
where $S_{V}$ is the set of validation examples. We will now provide a brief overview of the customary methods in the literature for the single level reduction of bilevel problems and their drawbacks in the context of hyperparameter optimization.

\subsection{First-order condition-based reduction}\label{subsec:KKT}

The first order conditions of the lower level problem can be utilized for the single level reduction of bilevel optimization problems. However, in the context of machine learning problems, especially the deep learning architectures, the lower level problem is usually highly non-linear with a multitude of local minimum, maximum and saddle points. Therefore, in such cases a first-order condition-based reduction would not be analogous to the original bilevel problem. In fact, such a formulation considers even the maximums and the saddle-points from the lower level as acceptable solutions at the upper level. The reduced formulation based on the first-order condition is as follows:
\begin{align*}
\minimize_{\lambda,w} \quad & F(w) \\
\st\quad  & \\
 & \hspace{-12mm} \nabla_{w} f(\lambda,w) = 0
\end{align*}
Even though the above formulation seems tractable, there are significant drawbacks to such reduction. Firstly, as mentioned above, the first-order condition accepts all the stationary points from the lower level problem. Secondly, solving the above formulation would require computing second-order derivatives, hence, it may not be feasible for a machine learning problem with a large number of model parameters. A study that uses this approach for hyperparameter optimization is \cite{mehra2019penalty}.

\subsection{Lower level reaction set mapping}\label{subsec:Psimapping}
Another method for single level reduction uses approximation of the reaction set mapping $\Psi:\reals^m\tos\reals^n$, that is given as follows:
\begin{align*}
\Psi(\lambda) = \argmin_{w}\{f(\lambda,w)\},
\end{align*}
The lower level problem is replaced by the above constraint, such that, $\Psi(\lambda)$ contains the optimal solutions (model parameters) of the lower level for any given $\lambda$. Mostly, the mapping isn't available; hence, it has to be approximated by solving multiple lower level problems corresponding to different values of $\lambda$. Since, the mapping is set-valued, the lower level problem may not contain unique optimal solutions for a given $\lambda$. 
The single-level reduction using $\Psi(\lambda)$ is expressed as a constrained optimization problem given below:
\begin{align*}
\minimize_{\lambda,w} \quad & F(w) \\
\st\quad  & \\
 & \hspace{-12mm} w \in \Psi(\lambda)
\end{align*}
The studies that rely on this mapping are \cite{lorraine2018stochastic,mackay2019self}. These studies make a strong assumption of uniqueness at the lower level, i.e. the lower level optimization problem is assumed to have a single optimum for any given $\lambda$. Such an assumption is often made because a set-valued mapping is hard to approximate and mathematically intractable.

\section{Proposed Method}\label{propmethod}
In this section, we propose a new and an alternative method for solving the hyperparameter optimization problem that does not suffer from the drawbacks of the previous approaches. Instead of aforementioned methods of single level reduction, our method approximates the lower level optimal value function mapping, known as $\varphi$-mapping. There are a number of studies that approximate the $\varphi$-mapping for solving bilevel optimization problems \cite{sinha2020bilevel,sinha2018bilevel}. The $\varphi$-mapping is a function, i.e. $\varphi: \mathbb{R}^{m} \to \mathbb{R}$, which is defined below:
\begin{align*}
\varphi(\lambda)=\minimize_w f(\lambda, w),
\end{align*}
This mapping represents the minimum lower level function value corresponding to any hyperparameter $\lambda$. Now, the bilevel hyperparameter optimization problem can be expressed as follows:
\begin{align}\label{mod:bilevelHO}
\begin{split}
\minimize_{\lambda,w} \quad & F(w) \\
\st\quad  & \\
 & \hspace{-12mm} f(\lambda, w) \le \varphi(\lambda)
\end{split}
\end{align}
Note that the constraint in the above formulation may be treated as an equality, which we will exploit later. The $\varphi$-mapping is usually not readily available, and hence, has to be approximated. The approximation is relatively simple, given that $\varphi(\lambda)$ is single-valued and a scalar. By solving the lower level problem for some upper level parameters, an approximation of this function can be made. In contrast to $\Psi$-mapping, $\varphi$-mapping avoids the complexities associated with approximating a set. 

\subsection{Approximating the $\varphi$-mapping}\label{varphiApprox}
In our study, we use Kriging for approximation of the $\varphi$-mapping. Kriging is a method for interpolation of a function, where the interpolation is governed by a Gaussian process. We generate a sample $\mathcal{S}$ of hyperparameters in the feasible region that are denoted as $\lambda^{(i)} : i \in \{1, \dots, L \}$. The lower level optimization problem is then solved for each $\lambda^{(i)}$ using the stochastic gradient method, thus providing the set of optimal lower level function values, i.e. $f(\lambda^{(i)}, w^{(i)}) = \min_{w} l(w, S^{T}) + \Theta(w,\lambda^{(i)})$. With the help of this sample data, we use Kriging to approximate the $\varphi$-mapping.
\begin{align*}
\varphi(\lambda^{(i)}) = \mu + \epsilon(\lambda^{(i)}) \quad i = 1, \ldots, L,
\end{align*}
where, $\mu$ is the expectation and $\epsilon(\lambda^{(i)}) \sim \mathcal{N}(0,\sigma^{2})$ is the error term. The errors are assumed to be negatively correlated with the measure of distance between them. The correlation between error terms is given as follows:
\begin{align*}
\mbox{Corr}[\epsilon(\lambda^{(i)}),\epsilon(\lambda^{(j)})] = e^{-d(\lambda^{(i)},\lambda^{(j)})},
\end{align*}
where $d(\lambda^{(i)},\lambda^{(j)}) = \Sigma_{k=1}^{n} \theta_k |\lambda_{k}^{(i)}-\lambda_{k}^{(j)}|^{p_k}$ represents weighted distance between two points $\lambda^{(i)}$ and $\lambda^{(j)}$. We can get the optimal value for the parameters by performing maximum likelihood estimation for the sample. For further details on Kriging, one can refer to \cite{sacks1989design,jones1998efficient}. The estimation provides us with the approximate mapping $\hat{\varphi}(\lambda)$, using which our single level hyperparameter optimization problem is approximately given as follows:
\begin{align}\label{mod:bilevelHOApprox}
\begin{split}
\minimize_{\lambda,w} \quad & F(w)\\
\st\quad  & \\
 & \hspace{-12mm} f(\lambda,w) \le \hat{\varphi}(\lambda)
\end{split}
\end{align}
Next, we discuss the method that is used to solve the above optimization problem.

\subsection{Augmented Lagrangian Method}\label{subsec:PenaltyMethod}
The problem stated in \ref{mod:bilevelHOApprox} is a non-linear constrained optimization problem that is converted into a penalized unconstrained optimization problem using the augmented Lagrangian method, with the assumption that the constraint holds with an equality, without any loss of generality. The unconstrained optimization problem can be solved using the stochastic gradient method with $w$ and $\lambda$ as variables.


\begin{definition}\label{def:bilevel6}
Let $P(\lambda, w) = (f(\lambda,w) - \varphi(\lambda))$, $R$ be the penalty parameter and $\mu$ be an estimate of the Lagrangian multiplier then the unconstrained problem to be solved is given as:
\begin{equation}\label{eq:unconstrained}
\minimize_{\lambda,w} Z(\lambda,w) = F(w) + \frac{R}{2} P(\lambda, w)^2 + \mu P(\lambda,w)
\end{equation}
where $R$ and $\mu$ are updated with an update rule in every iteration of unconstrained optimization, which leads to the optimum of problem~\ref{mod:bilevelHOApprox}.
\end{definition}
From here on we will refer to the augmented Lagrangian function $Z(\lambda,w)$ as the penalized loss function. The step-by-step procedure for our algorithm, which we refer to as the penalized validation method (PVM), is provided in Algorithm \ref{alg:pseudocode}.

\begin{algorithm*}[]
\small
\SetAlgoLined
\KwResult{$\lambda^{*}, w^{*}$}
 \textbf{Initialize:} Training set $S_T$, validation set $S_V$ and hyperparameter samples $\{\lambda^{(i)}\}_{i=1}^{L}$\;
    \For{$i = 1 \cdots L$}{
                $f(\lambda^{(i)}, w^{(i)}) = \min_{w} l(w, S^{T}) + \Theta(w,\lambda^{(i)})$\;
                $F(w^{(i)}) = l(w^{(i)}, S^{V})$
    }
    Obtain $\hat\varphi$-mapping using Kriging approximation between $\{f(\lambda^{(i)}, w^{(i)})\}_{i=1}^{L}$ and $\{\lambda^{(i)}\}_{i=1}^{L}$\;
    Let $v^0 =(\lambda^{(k)},w^{(k)})$, such that $F(w^{(k)}) = \minimize_i \{F(w^{(i)})\}_{i=1}^{L}$, penalty $R=R^0$ and multiplier $\mu = \mu^0$\;
    \For{$i=1 \cdots M$}{
            $\minimize_{(\lambda,w)} Z(\lambda,w) = F(w) + \frac{R^{i-1}}{2} (f(\lambda,w) - \varphi(\lambda))^{2} + \mu^{i-1} (f(\lambda,w) - \varphi(\lambda))$ with $v^{i-1}$ as starting point\;
            $(\lambda^{i}, w^{i}) \leftarrow \argmin_{(\lambda,w)} Z(\lambda,w)$\;
            $v^{i} \leftarrow (\lambda^{i}, w^{i})$\;
            $\mu^{i} = \mu^{i-1} + R^{i-1}(f(\lambda^i,w^i) - \varphi(\lambda^i))$\;
            $R^{i} = \eta R^{i-1}$\;
    }
    $(\lambda^{*}, w^{*}) \leftarrow v^{M}$
\caption{Pseudocode for the penalized validation method. Note that we use $R^0=2$, $\mu^0=2$ and $\eta=1.5$ in our experiments \cite{Nocedal2006}.}
\label{alg:pseudocode}
\end{algorithm*}

\section{Results}\label{sec:experiments}
 In this section, we provide the details of our experiments and discuss the results. To test our theoretical framework, we apply the proposed algorithm (PVM) to two datasets. The first dataset is the Communities and Crime (normalised) dataset \cite{Comm&Crim}, where the objective is to solve the ridge regression problem with a single hyperparameter to be optimized, and the second is the MNIST dataset \cite{lecun2010mnist}, where the objective is to solve a classification problem for which we construct 6 test cases with single and dual hyperparameters. 
 
 

\subsection{Communities and Crime Dataset}
\begin{figure}
\begin{minipage}{.49\textwidth}
\begin{center}
  \includegraphics[scale =0.49]{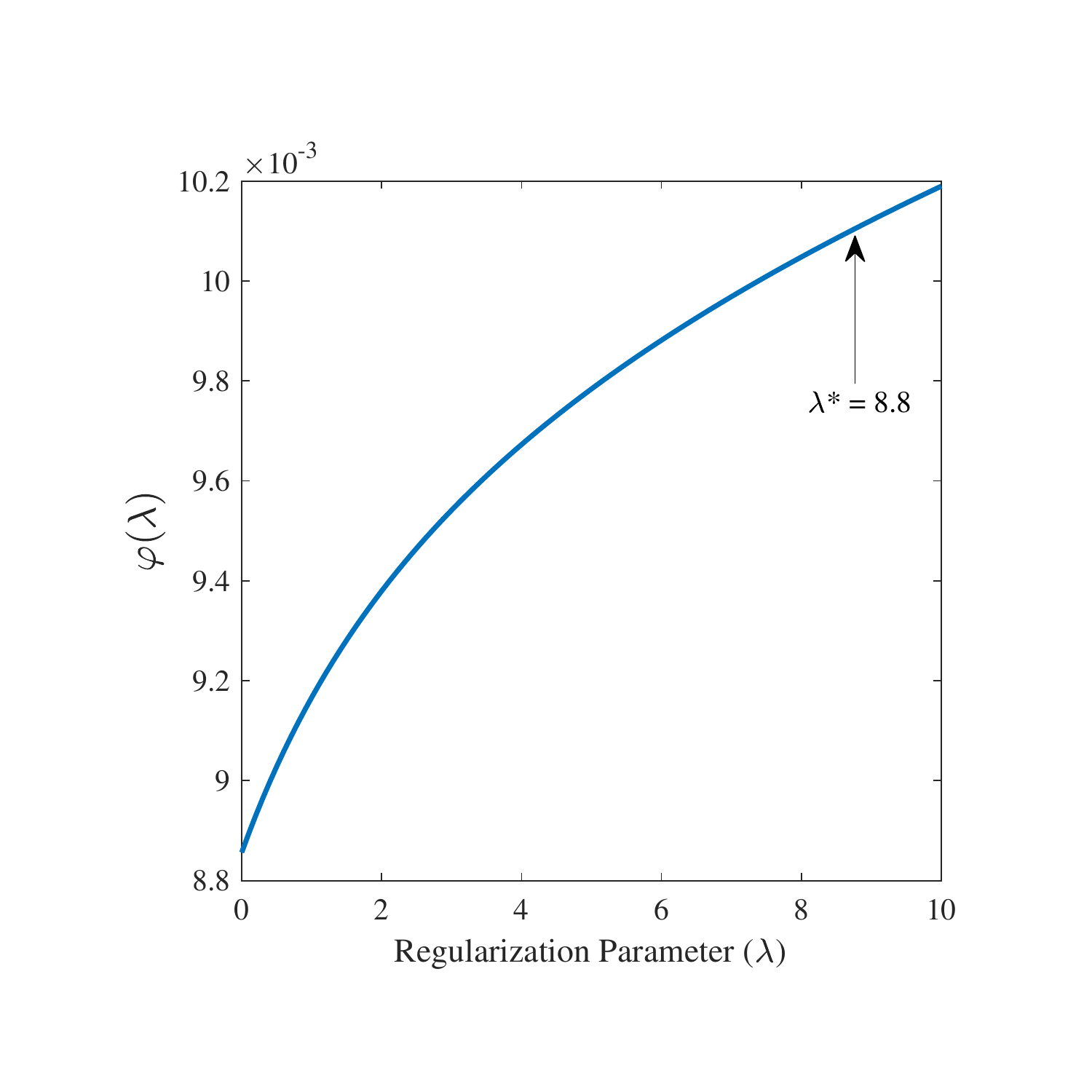}
  \captionof{figure}{$\varphi$-mapping for the communities and crime dataset obtained using grid search. The approximate $\hat{\varphi}$-mapping overlaps the actual mapping.}
  \label{fig:varphi}
  \end{center}
\end{minipage}\hfill
\begin{minipage}{.49\textwidth}
\begin{center}
  \includegraphics[scale =0.49]{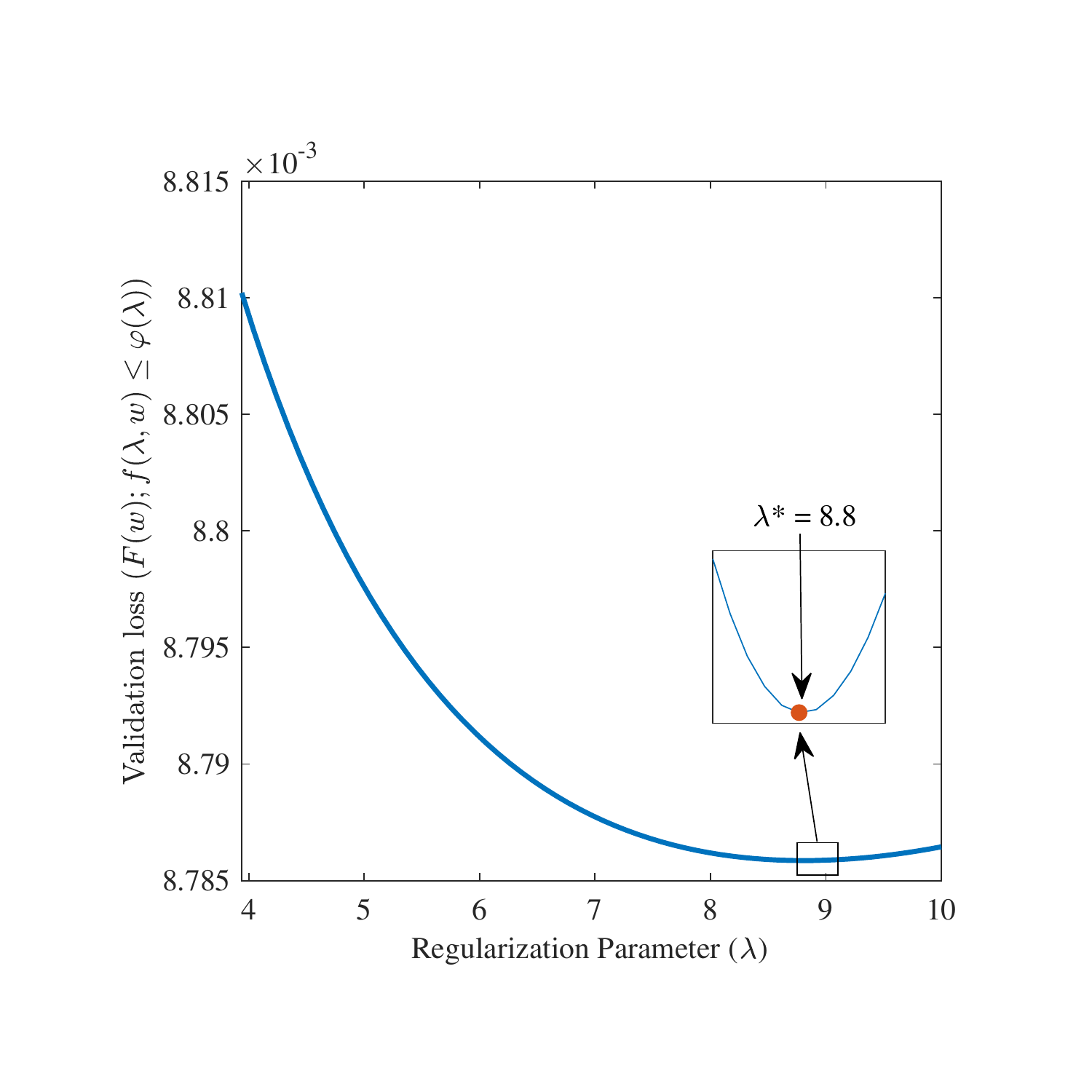}
  \captionof{figure}{Validation loss for communities and crime dataset over uniformly spaced hyperparameter grid. The optimum $\lambda^{*}$ found using PVM is shown.}
  \label{fig:validation}
  \end{center}
\end{minipage}
\end{figure}

The Communities and Crime dataset contains 122 independent variables and 1 dependent (continuous) variable, while the number of data points are 1994. The dataset contains some missing values for which we perform regression imputation. For this dataset, we use 55\% of the data for training, 20\% for validation and 25\% for testing. To identify the optimal $\lambda$ for the ridge regression problem, we sample 10 hyperparameters over a uniformly spaced grid, i.e. $\lambda \in [0,10]$, and identify the corresponding $\hat{\phi}$-mapping which is provided in Figure \ref{fig:varphi}. The approximate mapping ($\hat{\varphi}$) almost overlaps the true $\varphi$-mapping. Thereafter, we solve the augmented Lagrangian problem to arrive at the optimal $\lambda$, which is found to be $8.80$. We verify our solution using a grid of 100 points with $\lambda \in [0,9.90]$ for which the minimum validation loss occurs at $\lambda=8.80$ as shown in Figure \ref{fig:validation}. It is noteworthy that we are able to arrive at the optimal solution of $8.80$ using PVM by solving only 10 lower level optimization problems and the augmented Lagrangian optimization problem. We perform 4 iterations of augmented Lagrangian optimization, where each iteration is equal to 1 lower level optimization problem in terms of computational expense. Grid search requires 100 lower level optimization problems to be solved to arrive at the same answer. We also provide a comparison with Bayesian optimization\footnote{We use the sequential model based optimization (SMBO) algorithm as the Bayesian optimization technique.}, which also required close to 50 evaluations to approach the optimum. Additional details of our experiments for this dataset are provided in Table \ref{tab:results}.
\subsection{MNIST Dataset}
In the case of the MNIST dataset, which is a multi-classification problem, we use multi-layer perceptron (MLP) model with a single hidden layer consisting of 100 nodes. We create three datasets by random sampling from the original MNIST dataset, where each subset consists of 1000, 5000 and 10000 data points. For the MNIST experiments, we use 75\% of the sampled data points for training and 25\% for validation. The test set is a separate set that contains 10,000 points. We further divide the experiments into two parts, wherein the first set of experiments contain single regularization term that regularizes all the weight matrices of MLP, while the second set of experiments contain two regularization terms that regularize the two weight matrices of the MLP separately. Therefore, overall we conduct six experiments that we name as 1000 MNIST (1 HP), 5000 MNIST (1 HP), 10000 MNIST (1 HP), 1000 MNIST (2 HP), 5000 MNIST (2 HP) and 10000 MNIST (2 HP), where 1 HP and 2 HP denote the single and dual hyperparameter test cases. 

\begin{figure}
\centering
\begin{minipage}{.45\textwidth}
  \includegraphics[scale =0.49]{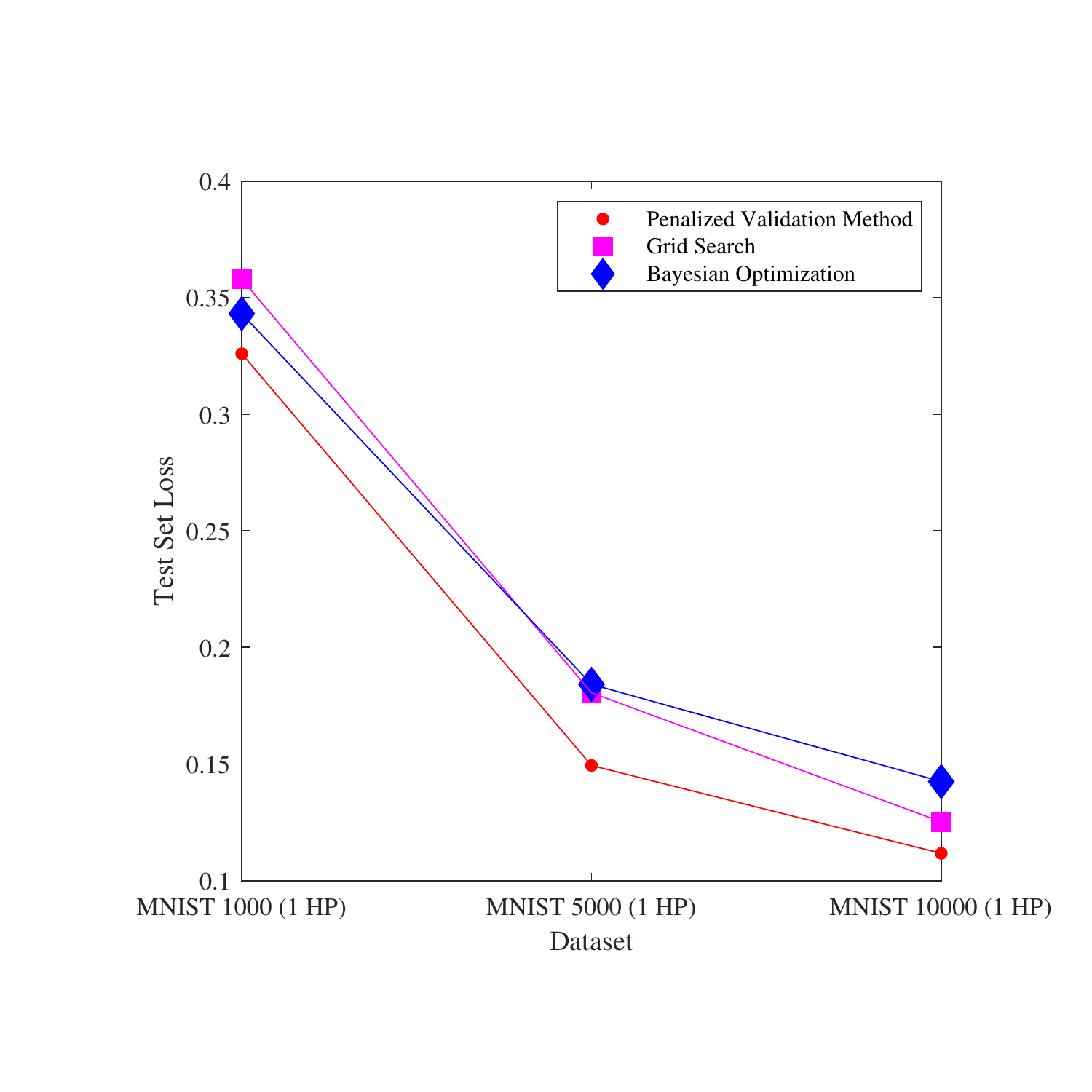}
  \vspace{-5mm}
  \captionof{figure}{Test loss performance of PVM, grid search and Bayesian optimization on 1 HP test cases.}
  \label{fig:comparision1}
\end{minipage}\hfill
\begin{minipage}{.45\textwidth}
  \includegraphics[scale =0.49]{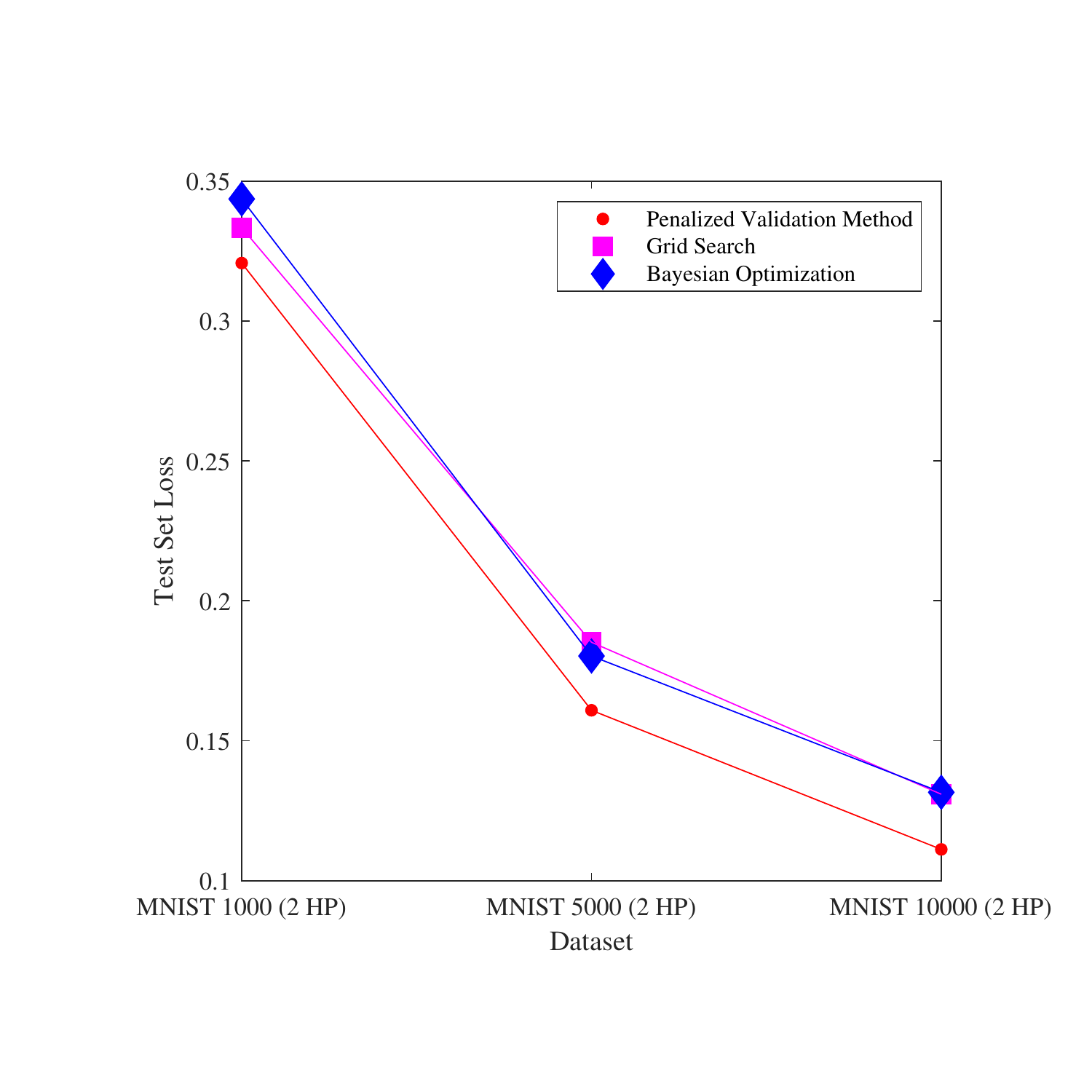}
  \vspace{-5mm}
  \captionof{figure}{Test loss performance of PVM, grid search and Bayesian optimization on 2 HP test cases.}
  \label{fig:comparision2}
\end{minipage}
\end{figure}

\begin{figure}
\centering
\begin{minipage}{.45\textwidth}
  \includegraphics[scale =0.49]{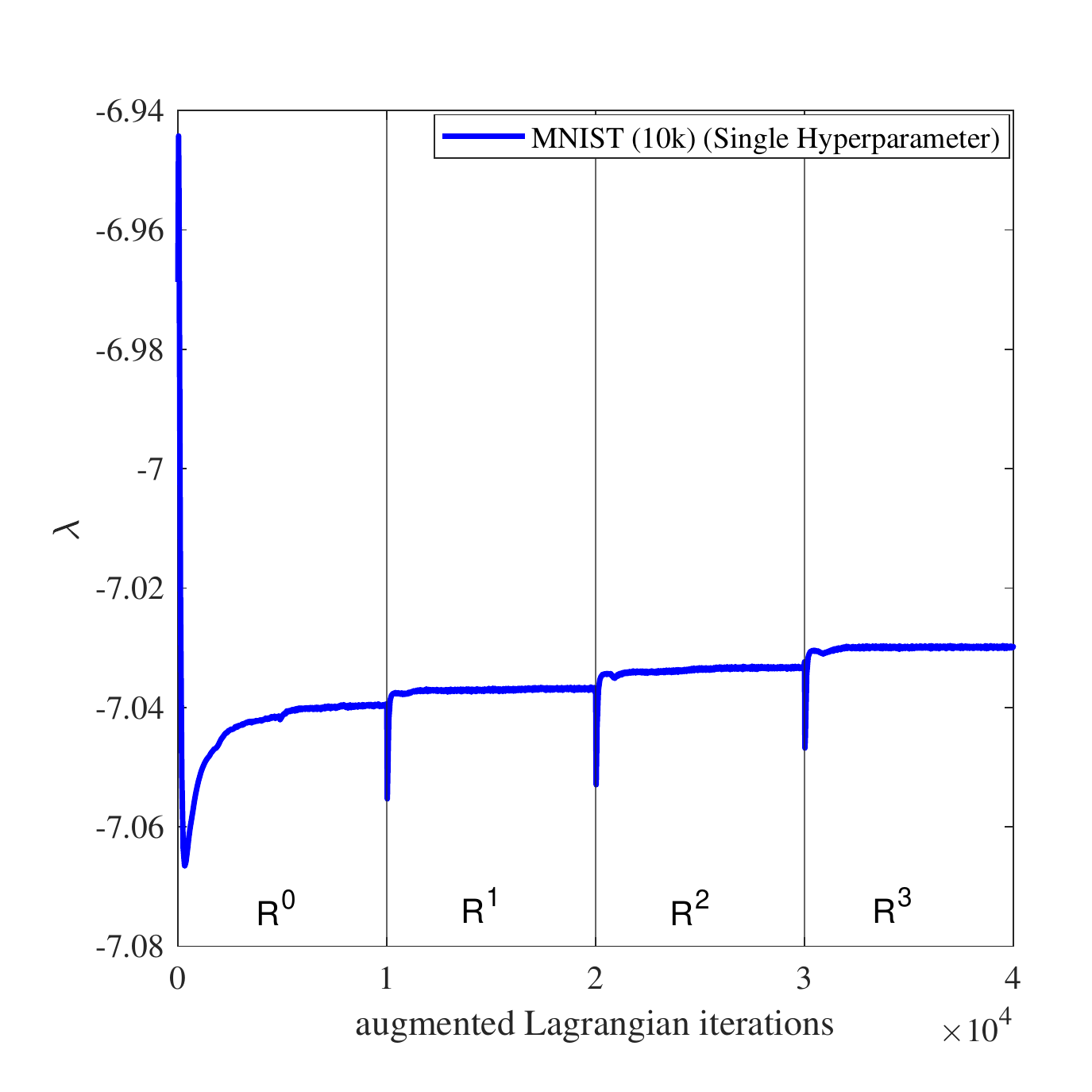}
  \captionof{figure}{Convergence plot for augmented Lagrangian optimization towards optimal $\lambda$ for the MNIST 10000 (1 HP) test case.}
  \label{fig:lambdaMovement1}
\end{minipage}\hfill
\begin{minipage}{.45\textwidth}
  \includegraphics[scale =0.49]{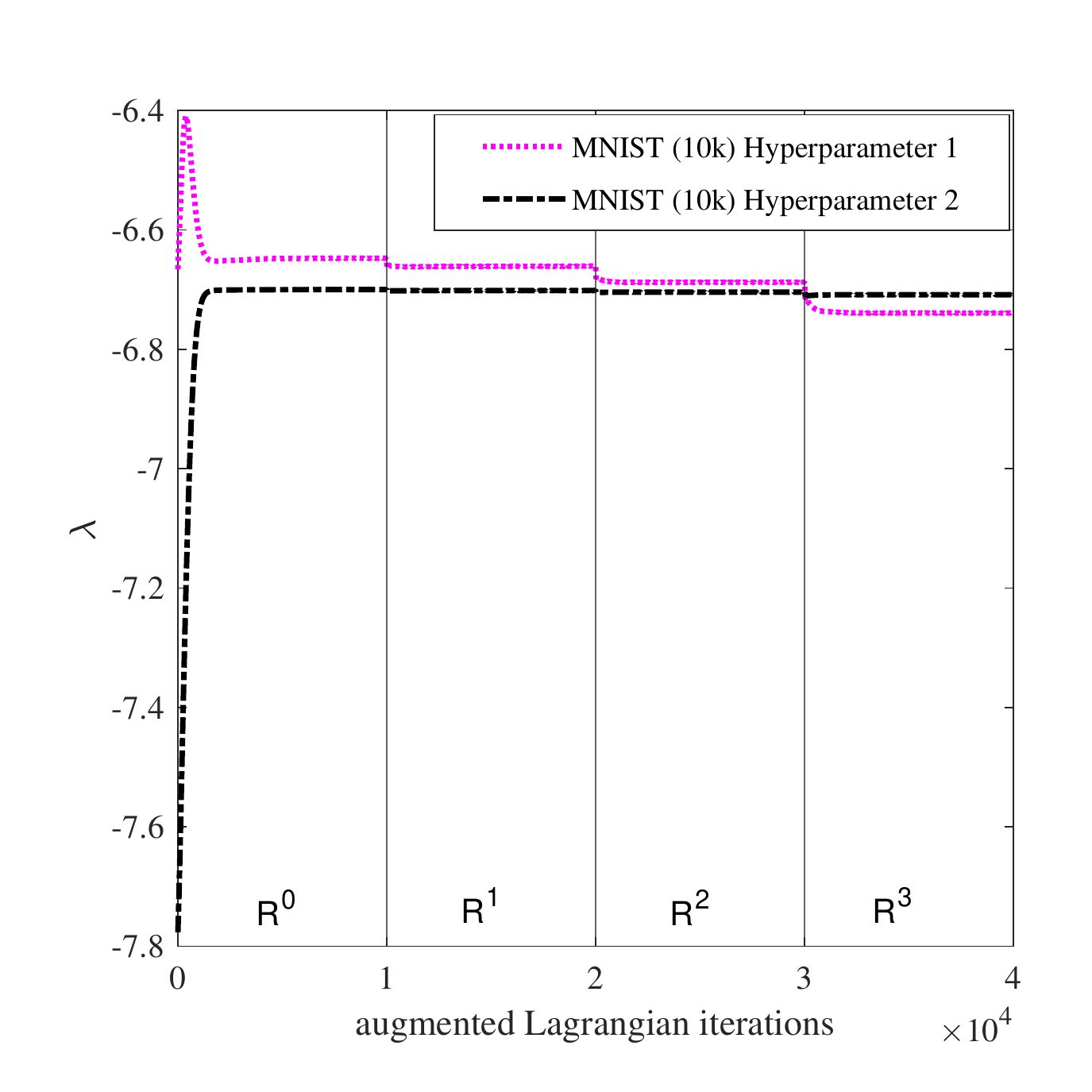}
  \captionof{figure}{Convergence plot for augmented Lagrangian optimization towards optimal $\lambda$ for the MNIST 10000 (2 HP) test case.}
  \label{fig:lambdaMovement2}
\end{minipage}
\end{figure}

In our experiments, we sample 10 hyperparameters ($\lambda = e^{\xi}: \xi \in [-10,0]$) with equally spaced values of $\xi$, and identify the corresponding $\hat{\phi}$-mapping, followed by the augmented Lagrangian optimization to minimize the penalized validation loss. Figures \ref{fig:comparision1} and \ref{fig:comparision2} provide a comparison of our approach against grid search and the Bayesian optimization method for the 1 HP and 2 HP cases, respectively. Clearly, our proposed method outperforms all the other techniques in terms of performance with respect to the test set loss for all the 6 test cases. Figures \ref{fig:lambdaMovement1} and \ref{fig:lambdaMovement2} provide the convergence plot of the augmented Lagrangian technique over iterations.

Table \ref{tab:results} provides a detailed summary of the experiments with different approaches for each test case. The computational expense of each method is measured in terms of the number of lower level optimization problems that have been solved (for different values of $\xi$ or $\lambda$). It is clear that the number of lower level optimization problems solved by our approach is significantly lower as compared to other methods and yet we achieve the best performance on the test set. Note that the proposed method (PVM) demands computational resources at two steps; firstly, while approximating the $\varphi$-mapping and secondly, while solving the augmented Lagrangian problem. We have accounted for both while reporting the computational expense of PVM.


\begin{table}[]
\begin{center}
\small
\caption{Results of the hyperparameter optimization problem on various dataset using PVM, grid search and Bayesian optimization. 1 HP denotes single hyperparameter test case and 2 HP denotes dual hyperparameter test case. Note that (+4) accounts for the additional computational requirements on account of augmented Lagrangian optimization.}
\label{tab:results}
\begin{tabular}{@{}lllllllll@{}}
\toprule
                                                                       &                                                                & \begin{tabular}[c]{@{}l@{}}Communities\\and\\Crime\end{tabular} & \begin{tabular}[c]{@{}l@{}}MNIST\\ (1000)\\(1 HP)\end{tabular} & \begin{tabular}[c]{@{}l@{}}MNIST\\ (5000)\\(1 HP)\end{tabular} & \begin{tabular}[c]{@{}l@{}}MNIST\\ (10000)\\(1 HP)\end{tabular} & \begin{tabular}[c]{@{}l@{}}MNIST\\ (1000)\\(2 HP)\end{tabular} & \begin{tabular}[c]{@{}l@{}}MNIST\\ (5000)\\(2 HP)\end{tabular} & \begin{tabular}[c]{@{}l@{}}MNIST\\ (10000)\\(2 HP)\end{tabular} \\ \midrule
\multirow{6}{*}{Bayesian}                                              & Training Loss                                                      & 0.0095                  & 0.0194                                                 & 0.0109                                                 & 0.0147                                                  & 0.0232                                                 & 0.0175                                                 & 0.0069                                                  \\\cmidrule{2-9}
                                                                       & Validation Loss                                                      & 0.0088                  & 0.2692                                                 & 0.2420                                                 & 0.1420                                                  & 0.2654                                                 & 0.2292                                                 & 0.1418                                                  \\\cmidrule{2-9}
                                                                       & Testing Loss                                                   & 0.0088                  & 0.3431                                                 & 0.1841                                                 & 0.1425                                                  & 0.3436                                                 & 0.1803                                                 & 0.1316                                                  \\\cmidrule{2-9}
                                                                       & \multirow{2}{*}{$\xi = \ln(\lambda)$}                                      & \multirow{2}{*}{$\ln(8.80)$} & \multirow{2}{*}{-6.16}                               & \multirow{2}{*}{-7.37}                               & \multirow{2}{*}{-7.68}                                & -6.88                                                & -6.89                                                & -9.99                                                \\
                                                                       &                                                                &                         &                                                        &                                                        &                                                         & -5.13                                                & -7.2                                                & -6.37                                                 \\\cmidrule{2-9}
                                                                       & \begin{tabular}[c]{@{}l@{}}Lower Level\\ Optimizations\end{tabular} & 50                     & 50                                                    & 100                                                    & 100                                                     & 1000                                                    & 1000                                                    & 1000                                                     \\\cmidrule(r){1-9}
\multirow{6}{*}{\begin{tabular}[c]{@{}l@{}}Grid\\ Search\end{tabular} }                                                  & Training Loss                                                     & 0.0095                  & 0.0067                                                 & 0.0170                                                 & 0.0120                                                  & 0.0215                                                 & 0.0141                                                 & 0.0177                                                  \\\cmidrule{2-9}
                                                                       & Validation Loss                                                    & 0.0088                  & 0.2628                                                 & 0.2440                                                 & 0.1375                                                  & 0.2536                                                 & 0.2086                                                 & 0.1340                                                  \\\cmidrule{2-9}
                                                                       & Testing Loss                                                   & 0.0088                  & 0.3579                                                 & 0.1806                                                 & 0.1252                                                  & 0.3333                                                 & 0.1852                                                 & 0.1309                                                  \\\cmidrule{2-9}
                                                                       & \multirow{2}{*}{$\xi = \ln(\lambda)$}                                      & \multirow{2}{*}{$\ln(8.80)$} & \multirow{2}{*}{-7.17}                               & \multirow{2}{*}{-6.97}                               & \multirow{2}{*}{-7.68}                                & -5.56                                                & -6.67                                                & -7.78                                                 \\
                                                                       &                                                                &                         &                                                        &                                                        &                                                         & -6.67                                                & -7.78                                                & -6.67                                                 \\\cmidrule{2-9}
                                                                       & \begin{tabular}[c]{@{}l@{}}Lower Level\\ Optimizations\end{tabular} & 100                     & 100                                                    & 100                                                    & 100                                                     & 1000                                                    & 1000                                                    & 1000                                                     \\\cmidrule(r){1-9}
\multirow{6}{*}{\begin{tabular}[c]{@{}l@{}}PVM\\\end{tabular}}   & Training Loss                                                      & 0.0095                  & 0.0105                                                 & 0.0127                                                 & 0.0179                                                  & 0.0194                                                 & 0.0169                                                 & 0.0234                                                  \\\cmidrule{2-9}
                                                                       & Validation Loss                                                  & 0.0088                  & 0.0180                                                 & 0.0119                                                 & 0.0150                                                  & 0.0310                                                 & 0.0119                                                 & 0.0156                                                  \\\cmidrule{2-9}
                                                                       & Testing Loss                                                   & 0.0088                  & 0.3260                                                 & 0.1494                                                 & 0.1117                                                  & 0.3207                                                 & 0.1609                                                 & 0.1112                                                  \\\cmidrule{2-9}
                                                                       & \multirow{2}{*}{$\xi = \ln(\lambda)$}                                      & \multirow{2}{*}{$\ln(8.80)$} & \multirow{2}{*}{-6.75}                               & \multirow{2}{*}{-7.05}                               & \multirow{2}{*}{-7.03}                                & -3.56                                                & -6.76                                                & -6.74                                                 \\
                                                                       &                                                                &                         &                                                        &                                                        &                                                         & -8.69                                               &-6.70                                                & -6.70                                                 \\\cmidrule{2-9}
                                                                       & \begin{tabular}[c]{@{}l@{}}Lower Level\\ Optimizations\end{tabular} & 10 (+4)                      & 10 (+4)                                                     & 10 (+4)                                                     & 10 (+4)                                                      & 50 (+4)                                                     & 50 (+4)                                                     & 50 (+4)                                                      \\\bottomrule
\end{tabular}
\end{center}
\end{table}

\section{Conclusions}\label{sec:conclusions}

In our work, we have proposed an algorithm for solving the hyperparameter optimization problem using a bilevel approach that relies on approximation of the $\varphi$-mapping. We first formulate the hyperparameter optimization problem as a bilevel optimization problem and then reduce it to a single level constrained problem, which is then solved using the augmented Lagrangian technique. The proposed algorithm is applied to two test problems for which the regularization hyperparameter(s) have been optimized along with the model parameters. Our experiments suggest that our method finds the optimal parameters for the ridge regression problem with little computational requirements. Our experiments on the MNIST dataset suggest that our method finds the best performing models for all the test cases, when compared with grid search and Bayesian optimization methods. The proposed method is a first-order gradient based method, thus being efficient in terms of both memory and time complexity. The increase in the number of hyperparameters would not be detrimental to the proposed technique, as it would only mean an increase in the sample size at the $\varphi$ mapping estimation step, and an increase of a few variables at the augmented Lagrangian optimization step.

\end{document}